\documentclass[preprint,12pt]{elsarticle}

\usepackage{amssymb}

\usepackage{hyperref} 
\usepackage{booktabs}

\begin{document}

\begin{frontmatter}



\title{Toward Zero-Shot Unsupervised Image-to-Image Translation}


\author[1,2]{Yuanqi Chen}

\author[1,2]{Xiaoming Yu}

\author[3]{Shan Liu}

\author[1,2]{Ge Li\corref{cor1}}
\ead{geli@ece.pku.edu.cn}

\address[1]{School of Electronic and Computer Engineering, Shenzhen Graduate School, Peking University, Shenzhen 518055, China}
\address[2]{Pengcheng Laborotory, Shenzhen 518055, China}
\address[3]{Tencent Inc., Shenzhen 518000, China}

\cortext[cor1]{Corresponding author}

\begin{abstract}
Recent studies have shown remarkable success in 
unsupervised image-to-image translation.
However,
if there has no access to enough images in target classes,
learning a mapping from source classes to the target classes always suffers from mode collapse,
which limits the application of the existing methods.
In this work,
we propose a zero-shot unsupervised image-to-image translation framework to address this limitation,
by associating categories with their side information like attributes.
To generalize the translator to previous unseen classes,
we introduce two strategies for exploiting the space spanned by the semantic attributes.
Specifically,
we propose to preserve semantic relations to the visual space
and expand attribute space by utilizing attribute vectors of unseen classes,
thus encourage the translator to explore the modes of unseen classes.
Quantitative and qualitative results on different datasets
demonstrate the effectiveness of our proposed approach.
Moreover,
we demonstrate that our framework can be applied to many tasks,
such as zero-shot classification and fashion design.
\end{abstract}




\end{frontmatter}


\def\EXP{\mathbb{E}}
\newcommand{\rec}[1]{\|{#1}\|_1}

\section{Introduction} \label{sec:introduction}



Deep learning models have achieved great success in image-to-image(I2I) translation tasks.
Recently,
there is a line of works aiming to learn mappings among multiple classes~\cite{choi2018stargan,yu2018singlegan,liu2018unified}. 
However,
the issue of class imbalance usually degrades the performance of these methods.
As for the facial attribute transfer task, samples of people with glasses are much smaller than those of people without glasses.
Then the image translation models tend to be biased towards the majority class and generate glasses with few styles~\cite{mariani2018bagan,ali2019mfc}.

\begin{figure}[t]
	\centering
	\includegraphics[width=0.8\linewidth]{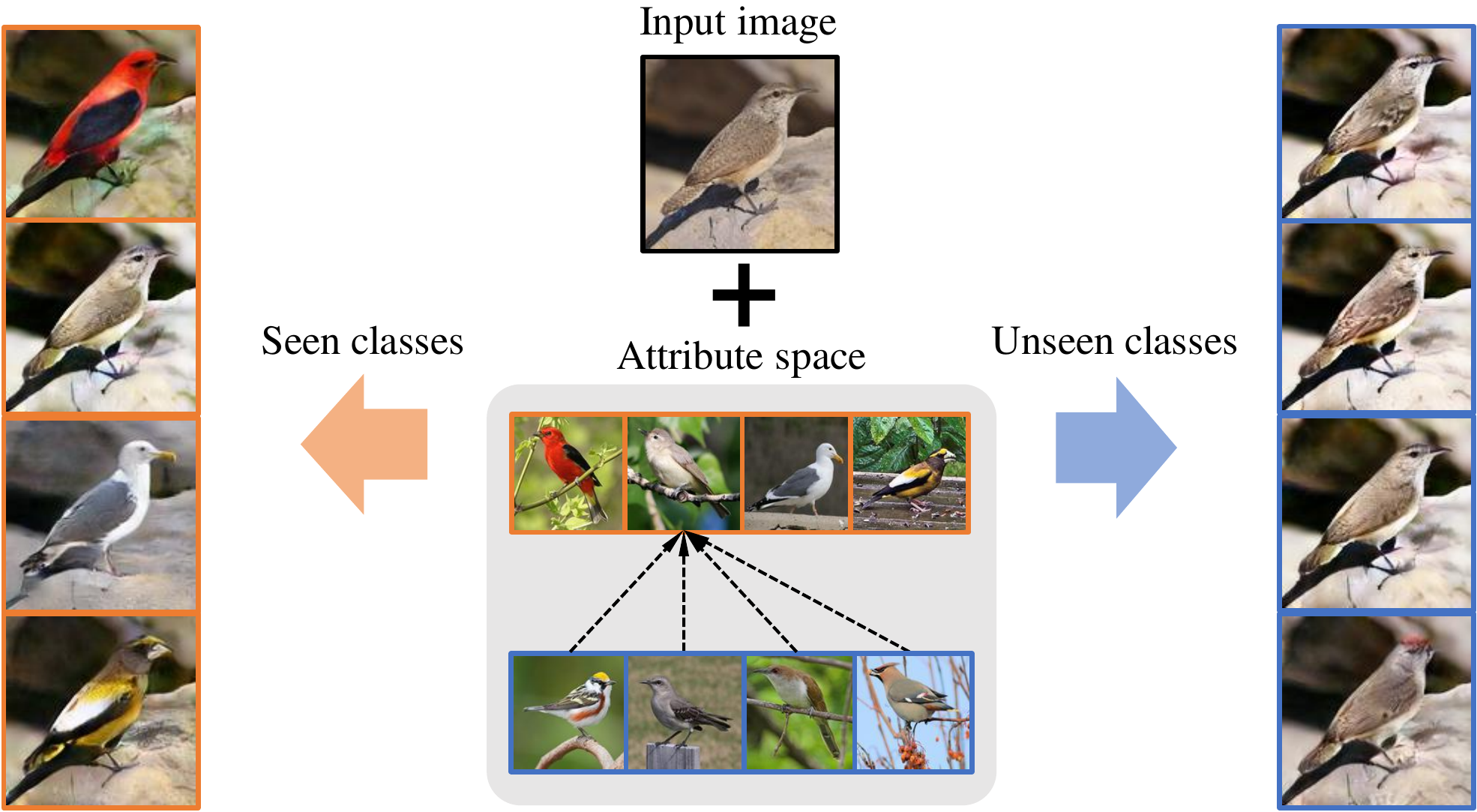}
	\caption{
	The limitation of existing image-to-image translation methods.
	They are successful in translating the input image to seen classes if these classes have enough samples during training time.
	However,
	when generating images of previous unseen classes,
	these methods suffer from the mode collapse problem,
	where the translator only produces outputs from a few modes of the data
	distribution.
	}
	\label{fig:seen_unseen}
\end{figure}

In this work,
we are going to explore the extreme case of the class imbalance issue, the zero-shot setting.
We begin by constructing a multi-domain I2I translation model as a baseline
to align the semantic attributes with the corresponding visual characteristics for seen classes.
However,
when translating the input image to unseen classes,
the translator suffers from the problem of mode collapse,
where the output images are usually collapsed to few modes specified by some seen classes.
As shown in Fig. \ref{fig:seen_unseen},
although we want to generate images of chestnut-sided warbler, 
mockingbird, 
black-billed cuckoo,
and Bohemian waxwing,
the four outputs all tend to have the same appearance as the warbling vireo,
which is one of the seen classes and has a similar appearance to the above four unseen classes.

Learning to map images from seen classes to unseen classes is challenging.
First,
paired training images are difficult to collect or even impossible.
For example,
it is hard to obtain an image pair of birds from different species,
but with the same pose and the same background.
Moreover,
the generalization capability is required for this task,
while the most existing techniques do not meet this requirement 
and suffer from the problem of mode collapse above.
Recently,
OST~\cite{benaim2018one} and FUNIT~\cite{liu2019few} are proposed to address these issues
in a one-shot manner and a few-shot manner respectively.
Different from these two methods,
we take a step further and focus on zero-shot unsupervised I2I translation.

To generalize the translator to unseen classes,
we hypothesize that by exploiting the space spanned by the semantic attributes,
the model can generate images of unseen classes conditioned on the corresponding attributes.
Specifically,
we propose to preserve semantic relations to the visual space
and expand attribute space by utilizing attribute vectors of unseen classes.
As pointed out in ~\cite{annadani2018preserving},
it is crucial to inherit the properties of semantic space for zero-shot learning.
To this end,
We first define the similarity metrics in the attribute space and the visual space separately,
then introduce a regularization term for preserving semantic relations.
As for expanding attribute space,
since attribute vectors of unseen classes have no corresponding images during training,
we propose an attribute regression loss for unseen semantic attribute vectors
to bridge the semantic attributes and the visual characteristics effectively.

Later in the experiments,
the effectiveness of our proposed method is evaluated on two datasets 
via quantitative metrics and qualitative comparisons.
We also show that the proposed method can be applied to many tasks,
such as zero-shot classification and fashion design.

The contributions of this work are summarized as follows:
\begin{itemize}
	\item We propose a framework for zero-shot unsupervised image-to-image translation,
	which alleviates the problem of mode collapse when synthesizing images of unseen classes.
	\item By preserving semantic relations and expanding attribute space, 
	our proposed model is efficient in utilizing the attribute space,
	and both quantitative and qualitative results on different datasets demonstrate its effectiveness.
	\item We demonstrate the application of our framework on the zero-shot classification task and fashion design task.
	We achieve competitive performance compared with existing methods.
\end{itemize}


\section{Related Work} \label{sec:related work}

\subsection{Image-to-Image Translation} 
aims to translate images from one visual domain to another.
Many computer vision tasks can be handled in image-to-image(I2I) translation framework,
e.g.,
image colorization~\cite{isola2017image}, 
image deblurring~\cite{lu2019unsupervised}, 
image super-resolution~\cite{ledig2017photo}, etc.
To learn convincing mappings across image domains from unpaired images,
CycleGAN~\cite{zhu2017unpaired}, DiscoGAN~\cite{kim2017learning} and DualGAN~\cite{yi2017dualgan}
introduce a cycle-consistency constraint and train two cross-domain translation models.
Recent works~\cite{choi2018stargan,yu2018singlegan,liu2018unified} extend the I2I framework from two domains to multiple domains under a single unified framework.
To learn an interpretable representation and further improve the model performance for I2I task,
there is a vast literature working on disentangling the representations
~\cite{liu2018unified,liu2019few,lample2017fader,huang2018multimodal,lee2018diverse,singh2019finegan,yu2019multi}.
MUNIT~\cite{huang2018multimodal} and DRIT~\cite{lee2018diverse} focus on disentangling the images into 
domain-invariant and domain-specific representations 
for producing diverse translation outputs.
FineGAN~\cite{singh2019finegan} disentangles the background, object shape, and object appearance to hierarchically generate images of fine-grained object categories.

However,
these methods fail to generalize to unseen domains based on prior knowledge of seen domains.
To improve the generalization capability of the I2I framework,
OST~\cite{benaim2018one} achieves one-shot cross-domain translation
using a single source class image and many target class images.
However,
OST can only map images between two classes.
By extracting appearance patterns from the target class images,
FUNIT~\cite{liu2019few} is capable to translate images of seen classes to analogous images of previously unseen classes.
Different from OST and FUNIT,
to further improve the machine imagination capability,
we assume that the images of target classes are unavailable even at test time
and explore zero-shot unsupervised image-to-image translation.

\subsection{Zero-Shot Classification}
aims to learn a model with generalization ability that can recognize unseen objects 
by only giving some semantic descriptions.
Since the seen objects and unseen ones are only connected in semantic space
and disjoint in the visual space,
early works~\cite{romera2015embarrassingly,socher2013zero,lampert2013attribute,frome2013devise,akata2015evaluation,akata2015label} learn a visual-semantic mapping with the seen samples.
During the test period,
unseen objects are projected from visual space to semantic space
and then classified by semantic attributes.
To address the hubness problem arising in the above methods,
~\cite{shigeto2015ridge,zhang2017learning} propose to use the visual space as the embedding space.
Another approach is to augment the training set with synthesized samples for unseen classes.
Several recent works~\cite{kumar2018generalized,xian2018feature,huang2019generative,li2019leveraging,elhoseiny2019creativity}
focus on generating new visual features conditioned on semantic attributes of novel classes.
With the synthesized data of unseen classes,
the zero-shot classification becomes a conventional classification problem.
There are also some works~\cite{josephzero,kim2020unseen} that try to generate images of unseen classes rather than feature level synthesis.
Different from these methods,
our goal is to translate images of seen classes to unseen classes while remaining class-independent information.


\section{Proposed Method} \label{sec:proposed method}

The definition of zero-shot image-to-image translation is as follows.
Let $ \mathcal{S} = \left\{ (x, y, a) | x \in \mathcal{X}^s, y \in \mathcal{Y}^s, a \in \mathcal{A}^s \right\} $ 
be a set of seen samples consisting of 
image $x$,
class label $y$,
and semantic attribute $a$.
$a$ is the class embedding of class $y$ that models the semantic relationship between classes.
As for unseen classes,
we have no access to the unseen image set $\mathcal{X}^u$ at training time,
and an auxiliary training set is 
$ \mathcal{U} = \left\{ (y, a) | y \in \mathcal{Y}^u, a \in \mathcal{A}^u \right\} $.
In zero-shot learning setting,
the seen and the unseen classes are disjoint, 
i.e., 
$ \mathcal{Y}^s \cap \mathcal{Y}^u = \emptyset $.
The goal of zero-shot image-to-image translation is to learn a mapping 
$ f : \mathcal{X}^s \times (\mathcal{A}^s \cup \mathcal{A}^u) \rightarrow \mathcal{X}^s \cup \mathcal{X}^u $,
which indicates that the translator should be capable to conduct the mapping for both seen and unseen classes.

We begin by constructing a multi-domain I2I translation model
as a baseline in Section \ref{subsec:multi-domain I2I},
and then introduce the proposed strategies that enable generalizing to unseen classes
in Section \ref{subsec:zero-shot I2I}.

\subsection{Multi-Domain I2I Translation} \label{subsec:multi-domain I2I}

We first consider how to learn a mapping
$ f^s : \mathcal{X}^s \times \mathcal{A}^s \rightarrow \mathcal{X}^s $
and build a multi-domain I2I translation model as a baseline.
Although lacking the generalization capability to unseen classes,
this baseline model learns to 
align the semantic attributes with the corresponding visual characteristics,
e.g.,
learns the relationship between the attribute "white wings" and its visual representations.

The proposed baseline model consists of a conditional generator $G$
and a multi-task discriminator $D$.
$G$ learns to translate an input image $x_i$ to an output image $x_t$
conditioned on the target attribute vector $a_t$,
i.e., $ G(x_i, a_t) \rightarrow x_t $.
It is noteworthy that $x_i$ and $a_t$ are both sampled from the set of seen samples $\mathcal{S}$.
To make full use the information in the semantic attribute $a_t$,
each residual block in the decoder of $G$ is equipped with 
adaptive instance normalization(AdaIN)~\cite{huang2017arbitrary,huang2018multimodal} for information injection.
As for the discriminator $D$,
we equip it with an auxiliary attribute regressor $R$~\cite{zhu2017toward,chen2016infogan}
to discriminate whether the output image $x_t$ has the visual characteristics of the conditional attribute $a_t$.

\subsubsection{Adversarial Loss.}
To match the distribution of synthesized images to the real data of the corresponding class,
we adopt an adversarial loss
\begin{equation}
\mathcal{L}_{adv} = \EXP_{(x_i, y_i) \sim \mathcal{S}} [\log(D(x_i, y_i))] + \EXP_{x_i, (a_t, y_t) \sim \mathcal{S}} [\log(1 - D(G(x_i,a_t), y_t))],
\end{equation}
where $D$ tries to distinguish between real and generated images of the given class label,
while $G$ attempts to generate realistic images. 

\subsubsection{Attribute Regression Loss.}
To encourage the generator $G$ to utilize the attribute vector,
we introduce an attribute regression by adding an auxiliary regressor $R$ on top of $D$.
The attribute regression loss of generated images can be written as
\begin{equation}
\mathcal{L}_{attr}^f = \EXP_{x_i, a_t \sim \mathcal{S}} [-\log(R(a_t | G(x_i,a_t)))],
\end{equation}
where $R(a_t | G(x_i,a_t))$ approximates the posterior $P(a_t | G(x_i,a_t))$.
On the other hand,
the attribute regression loss of real images is defined as
\begin{equation}
\mathcal{L}_{attr}^r = \EXP_{(x_i, a_i) \sim \mathcal{S}} [-\log(R(a_i | x_i))].
\end{equation}


\subsubsection{Self-Reconstruction Loss.}
In addition to the above losses,
we impose a self-reconstruction loss to facilitate the training process,
which can be written as 
\begin{equation}
\mathcal{L}_{rec} = \EXP_{(x_i, a_i) \sim \mathcal{S}} [\rec{G(x_i, a_i) - x_i}].
\end{equation}

\begin{figure*}[t]
	\centering
	\includegraphics[width=1\linewidth]{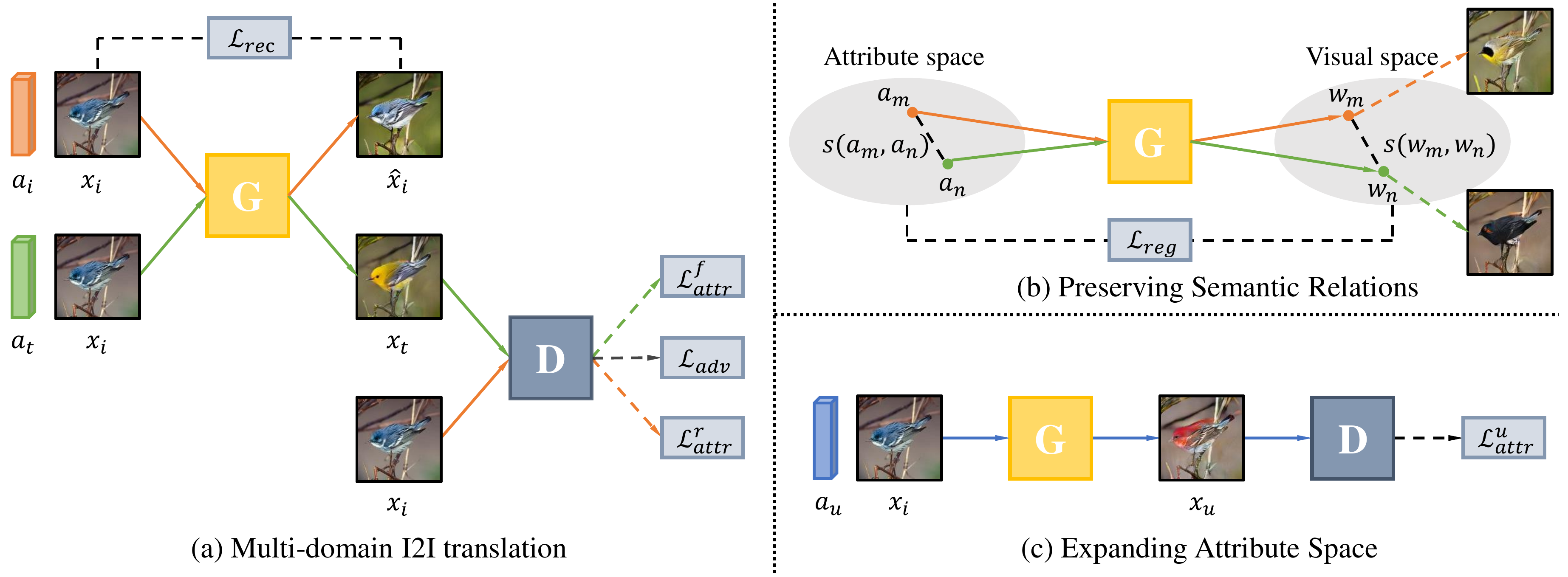}
	\caption{\textbf{Overview.}
	(a) The multi-domain I2I translation model learns to align the semantic attributes with the corresponding visual characteristics for seen classes.
	(b) To preserve semantic relations,
	for the sampled classes $y_m$ and $y_n$,
	we first calculate their similarity in the attribute space and visual space, 
	then we regularize them to be close to each other.
	(c) To expand attribute space,
	we utilize the attribute $a_u$ sampled from unseen classes and
	introduce an attribute regression loss for effective incentives
	to synthesize images with the visual characteristics of $a_u$.
	}
	\label{fig:framework}
\end{figure*}

\subsection{Generalizing to Zero-Shot I2I Translation} \label{subsec:zero-shot I2I}

Recall that our goal is to learn a mapping 
$ f : \mathcal{X}^s \times (\mathcal{A}^s \cup \mathcal{A}^u) \rightarrow \mathcal{X}^s \cup \mathcal{X}^u $.
With the above multi-domain I2I translator,
we can conduct the mapping 
$ f^s : \mathcal{X}^s \times \mathcal{A}^s \rightarrow \mathcal{X}^s $
and synthesize images of seen classes with high fidelity.
However,
when it comes to unseen classes,
the translator suffers from mode collapse
~\cite{goodfellow2014generative,yang2018diversity,mao2019mode}.
As shown in Fig. \ref{fig:seen_unseen},
although the attribute vectors from different unseen classes are injected,
the output images are collapsed to few modes specified by some seen classes,
which indicates that the translator tends to ignore some information in the attribute vectors of unseen classes.
we hypothesize that by exploiting the space spanned by the semantic attributes,
the model can generate images of unseen classes conditioned on the corresponding attributes.


\subsubsection{Preserving Semantic Relations.}
For unseen classes,
as there are no images available during the training process,
we need to use the attribute vectors to provide effective guidance on where to map the input image.
To this end,
we propose to preserve the relations in the attribute space to the visual space.
We begin by defining the similarity metrics in the attribute space and visual space separately,
for any two sampled classes $y_m$ and $y_n$.
For the relations in the attribute space,
We define the semantic relationship measure of two attribute vectors $a_m$ and $a_n$ 
using cosine similarity ~\cite{annadani2018preserving}: 
\begin{equation}
s(a_m, a_n) = \frac{<a_m, a_n>}{\|a_m\|_2 \|a_n\|_2}.
\end{equation}

However,
due to the high dimensionality of images,
it's hard to measure the relationship in the visual space.
To this end,
we propose a simple yet effective approach to inherit the relations of the attribute space 
to the visual space.
As our proposed generator is style-based,
given the input image $x_i$ and the target attribute $a_t$,
AdaIN~\cite{huang2017arbitrary,huang2018multimodal} first learns a set of affine transformation parameters $w_t = (\gamma_t, \beta_t)$ computed using $a_t$ via a multi-layer fully connected network.
It then adjusts the activations of $x_i$ to fit the data distribution specified by $a_t$
\begin{equation}
AdaIN(h_k, w_t) = \gamma_{t,k}\frac{h_k - \mu(h_k)}{\sigma(h_k)} + \beta_{t,k}, 
\end{equation}
where each feature map $h_k$ of $x_i$ is normalized separately,
and then scaled and biased using the corresponding scalar components from $w_t$.
In other words,
for any two sampled classes $y_m$ and $y_n$,
their learned affine transformation parameters $w_m$ and $w_n$ 
describe the characteristics of their data distributions in visual space. 
In this way,
we can utilize the similarity between $w_m$ and $w_n$
to approximate the similarity between the visual subspaces of $y_m$ and $y_n$
\begin{equation}
s(w_m, w_n) = \frac{<w_m, w_n>}{\|w_m\|_2 \|w_n\|_2}.
\end{equation}
We then introduce a regularization term to preserve the relations in the attribute space to the visual space
\begin{equation}
\mathcal{L}_{reg} = \EXP_{a_m, a_n \sim \mathcal{S}} [\| s(a_m, a_n) - s(w_m, w_n) \|_2].
\end{equation}

The proposed strategy shares some similarities to DistanceGAN~\cite{benaim2017one}, 
DSGAN~\cite{yang2018diversity},
and MSGAN~\cite{mao2019mode}.
DistanceGAN learns the mapping between the source domain and the target domain in a one-sided unsupervised way, 
by preserving the distance between two samples in the source visual domain to the target visual domain.
Different from it,
we preserve the relations in the attribute space to the visual space. 
To produce diverse outputs,
DSGAN and MSGAN propose to maximize the distance between generated images with respect to 
that between the corresponding latent codes.
Instead of maximizing the ratio of the distances in the visual space and the latent space,
our proposed strategy encourages the generator to maintain this ratio to 1 for aligning the semantic attributes with the corresponding visual characteristics.
Moreover,
we calculate the distance in the visual space using high-level statistics rather than raw RGB image values,
which alleviates the bias caused by some class-invariant information such as pose and lighting.


\subsubsection{Expanding Attribute Space.}

The mode collapse problem is similar to the phenomenon 
in imbalanced learning~\cite{kim2020unseen}/zero-shot classification ~\cite{chao2016empirical,song2018transductive}.
There exists a strong mapping bias during the phase of 
bridging the semantic attributes and the visual characteristics,
which causes degradation in performance when generating images of unseen classes.
In the training phase of the above multi-domain I2I translation model,
as we only see samples from seen classes,
the generator tends to map the input image to some visual subspaces 
specified by the seen classes.
In this way,
given a novel semantic attribute $a_u$ from an unseen class $y_u$,
the output image,
although realistic,
tends to own visual characteristics of some seen classes rather than the given class $y_u$.

To address the above problem,
we explore the use of the semantic attributes in $\mathcal{A}_u$ to expand attribute space.
Since the novel attribute $a_u$ has no corresponding images during training,
the above adversarial training can not be applied in this case.
For effective incentives to synthesize images with the visual characteristics of $a_u$,
we propose an attribute regression loss for unseen semantic attributes: 
\begin{equation}
\mathcal{L}_{attr}^u = \EXP_{x_i \sim \mathcal{S}, a_u \sim \mathcal{U}} [-\log(R(a_u | G(x_i,a_u)))],
\end{equation}
where $R(a_u | G(x_i,a_u))$ approximates the posterior $P(a_u | G(x_i,a_u))$.
Similar to InfoGAN~\cite{chen2016infogan},
the above loss term maximizes the mutual information between the attribute vector $a_u$ and the generated image $G(x_i,a_u)$. 
In this way,
the generator $G$ has access to the semantic attributes in $\mathcal{A}_u$ during the training phase,
so these semantic attributes are not novel for $G$ during the testing phase,
which helps to alleviate the mapping bias for unseen classes.

Finally,
the objective functions to optimize $G$ and $D$ are written as
\begin{equation}
\mathcal{L}_{D} = -\mathcal{L}_{adv} + \lambda_{cls} \mathcal{L}_{attr}^r,
\end{equation}
\begin{equation}
\mathcal{L}_{G} = \mathcal{L}_{adv} + \lambda_{cls} \mathcal{L}_{attr}^f + \lambda_{rec} \mathcal{L}_{rec} + \lambda_{reg} \mathcal{L}_{reg} + \lambda_{cls} \mathcal{L}_{attr}^u,
\end{equation}
where the hyper-parameters $\lambda_{cls}$, $\lambda_{rec}$, and $\lambda_{reg}$ 
control the importance of the corresponding term.


\section{Experiments} \label{sec:experiments}

\subsection{Implementation Details}

For generator $G$,
it contains an encoder-decoder architecture and 
each residual block in the decoder of $G$ is equipped with 
adaptive instance normalization(AdaIN)~\cite{huang2017arbitrary,huang2018multimodal} for information injection.
For discriminator $D$,
we adopt the architecture of multi-task adversarial discriminator in FUNIT~\cite{liu2019few},
which leverages PatchGANs~\cite{isola2017image} to classify whether local image patches are real or fake conditioned on the given class label.
We build our model on the hinge variant of GANs~\cite{miyato2018spectral,zhang2019self},
which uses a hinge loss to train the model instead of a cross-entropy loss. 
We use the real gradient penalty regularization~\cite{mescheder2018training} to stabilize the training process.
The hyper-parameters are set to be: 
$\lambda_{cls}=1$,
$\lambda_{rec}=0.1$,
and $\lambda_{reg}=10$.
We train all our models with Adam optimizer~\cite{kingma2014adam} with the learning rate of 0.0001 
and exponential decay rates $(\beta_1, \beta_2) = (0.5,0.999)$
on a single NVIDIA V100 GPU.
We refer the reader to our supplementary materials for more details about the network architecture.

\subsection{Datasets}

\subsubsection{CUB.} 
Caltech-UCSD Birds-200-2011(CUB)~\cite{wah2011caltech} dataset consists of 11,788 images
which come from 200 bird species.
312 attribute labels are perceived by MTurkers for each image.
The training set and test set of CUB have 150 and 50 species,
respectively.

\subsubsection{FLO.} 
Oxford Flowers(FLO)~\cite{nilsback2008automated} is another dataset commonly used in the zero-shot learning tasks.
It contains 102 flower categories with 8,189 images.
For this dataset,
we use the text embeddings as side information provided by~\cite{reed2016learning}.
FLO is split into 82 training classes and 20 test classes.

\subsection{Metrics}

\subsubsection{Fréchet Inception Distance.}
To quantify the performance,
we adopt Fréchet Inception Distance (FID)~\cite{heusel2017gans}
to evaluate the quality of generated images.
For each unseen class,
we first randomly sample images from the seen classes as input images 
with the number similar to the number of real images in this unseen class.
Then we synthesize novel images with the semantic attribute of the class based on these sampled images.
After synthesis,
we compute the FID score between the distribution of generated images and real images of all unseen classes.
For seen classes,
we conduct the same operation and report the FID score 
to evaluate the performance of seen classes.
The lower the FID score, the better the quality of generated images.

\subsubsection{Classification Accuracy.}
To measure whether a translation output belongs to the target class,
following FUNIT~\cite{liu2019few},
we adopt an Inception-V3~\cite{szegedy2016rethinking} classifier which is trained using all the classes of the dataset.
We report both Top-1 and Top-5 accuracies for unseen classes and seen classes.
The higher the accuracy,
the more relevant the generated images are to their translated class.
We also report these two metrics for ground truth(GT) in the test set.

\subsubsection{Human Perception.}
To judge the visual realism of the generated images,
we perform a user study on Amazon Mechanical Turk (AMT) platform.
For each task,
we randomly generate 2,500 questions by sampling images of all unseen classes.
The workers are given four target class images and a series of translation outputs generated by different methods.
They are given unlimited time to choose which synthesis is more similar to the target class images.
Each question is answered by 5 different workers.


\subsection{Main Results}

Since there is no previous unsupervised I2I translation method that is designed for our setting,
we adopt FUNIT~\cite{liu2019few} that is the most related state-of-the-art method for few-shot unsupervised I2I translation.
At test time,
given few images from the target category,
FUNIT extracts appearance patterns from them and applies these patterns to input images for the translation task,
while we assume that the images of the target category are unavailable even at test time.
We evaluate its performance under the 1-shot and 5-shot settings
and denote as FUNIT-1 and FUNIT-5,
respectively.
As for fair comparison,
we compare our proposed framework against StarGAN~\cite{choi2018stargan},
which is the state of the art for multi-class unsupervised I2I translation.

\subsubsection{Qualitative Evaluation.}

\begin{figure*}[t]
	\centering
	\includegraphics[width=1\linewidth]{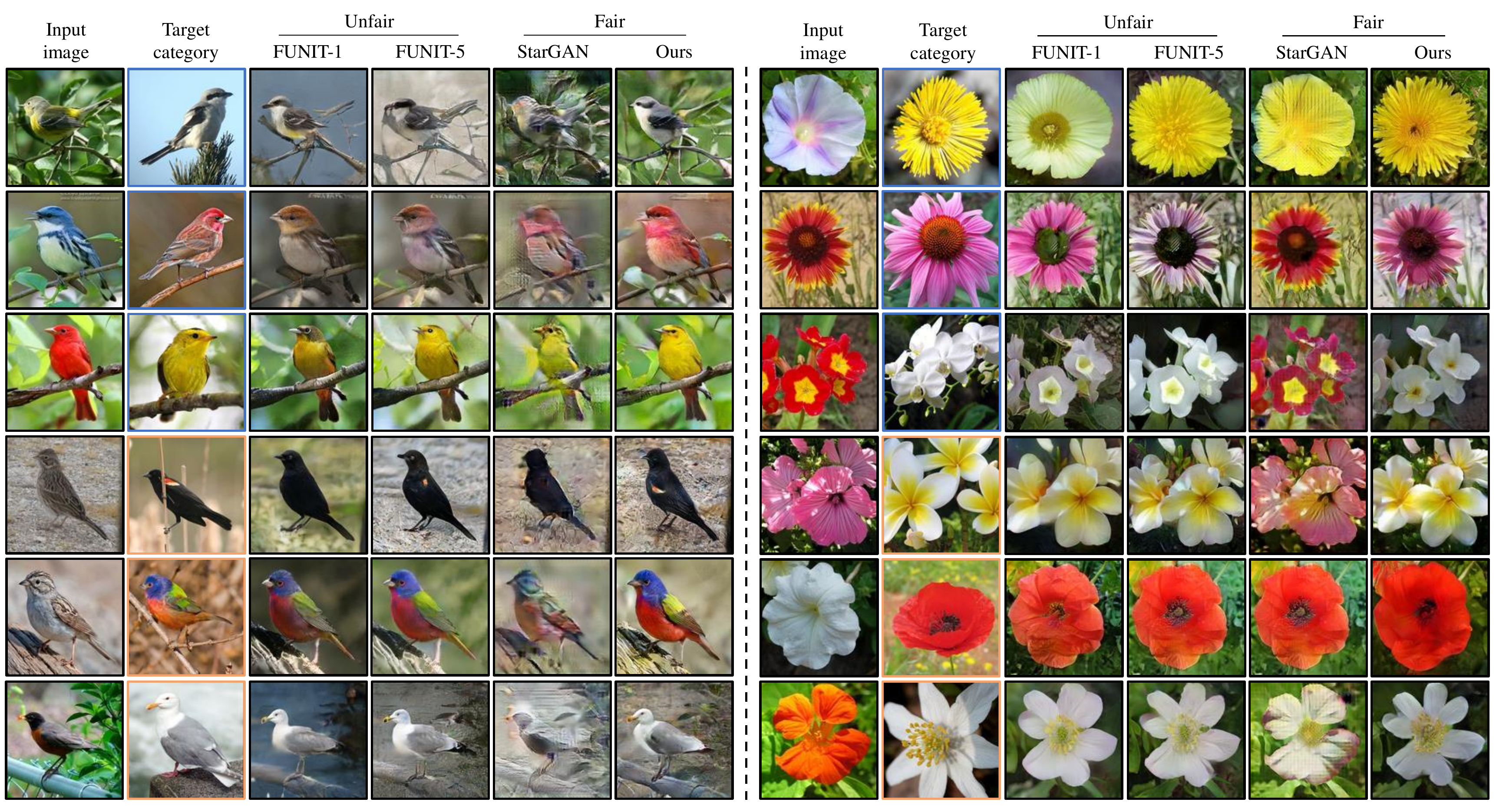}
	\caption{
	Qualitative comparison on CUB and FLO.
	For each dataset,
	the first column shows the input images sampled from seen classes,
	and the second column represents the characteristics of the target category.
	Each of the remaining columns shows the outputs from a method.
	The blue and orange borders indicate that the target category is sampled from unseen classes and seen classes,
	respectively.
	}
	\label{fig:examples}
\end{figure*}

As the qualitative comparison in Fig. \ref{fig:examples} shows,
the outputs of FUNIT-1 and FUNIT-5,
although realistic, 
can not well preserve the background of the input images,
which is caused by their example-guided image translation.
During the phase of extracting appearance patterns from example images,
the model would extract not only class-specific characteristics of the target category,
but also some other information in the example images,
e.g. background.
The more example images they have,
the more characteristics of the target category they capture.
In this way,
the outputs of FUNIT-5 tend to be more relevant to their target categories than the outputs of FUNIT-1.
As for fair comparison,
we observe that StarGAN produces results with significant artifacts
and can not well obtain the visual characteristics of the target class from its semantic attribute.
We conjecture that this is because StarGAN lacks enough incentives to utilize the attribute vector.
In contrast,
for both unseen and seen classes,
our method can generate realistic images while relevant to their target categories.

\begin{figure}[t]
	\centering
	\includegraphics[width=0.95\linewidth]{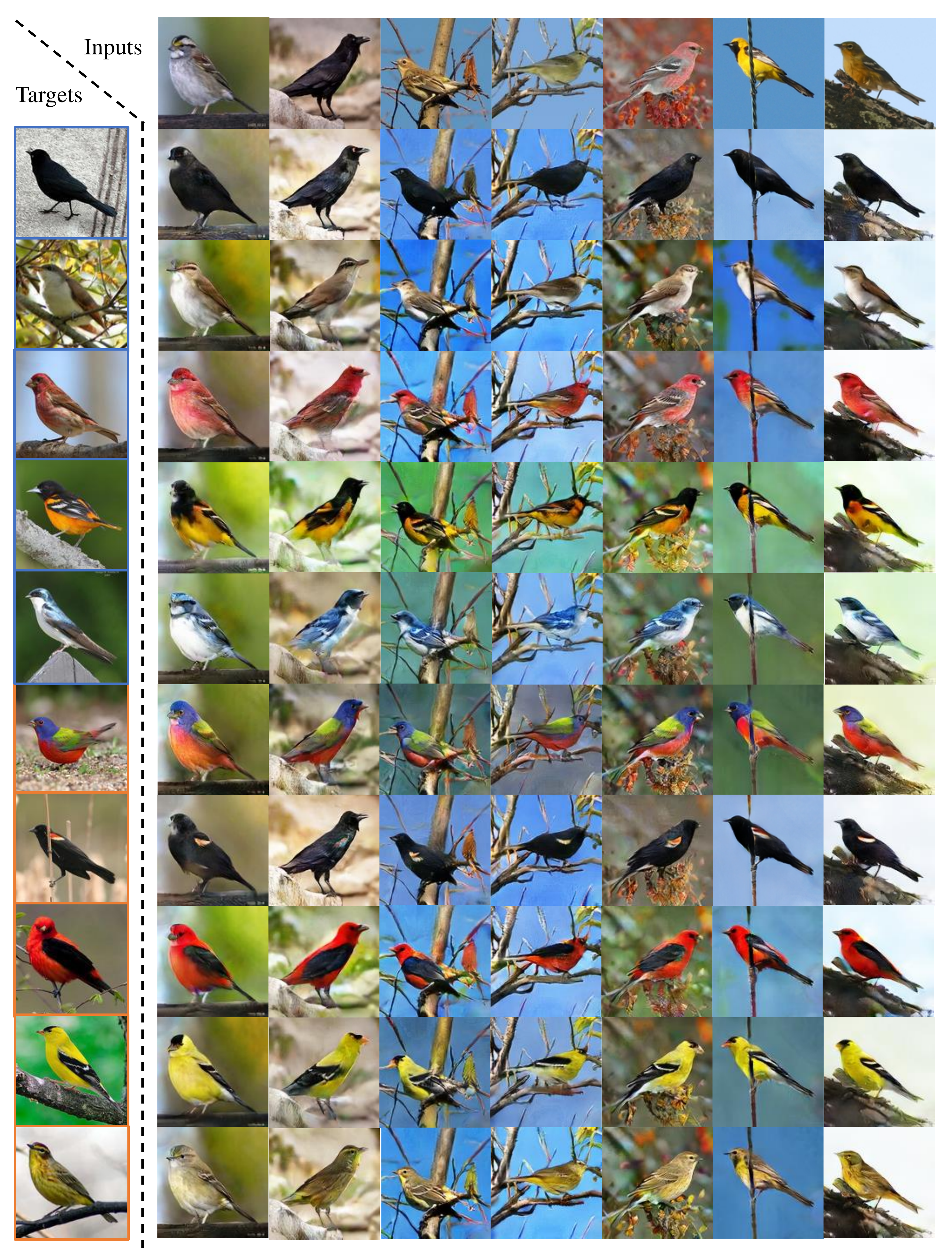}
	\caption{
	\textbf{Qualitative results on CUB.} 
	The first row shows the input images sampled from seen classes.
	The first column represents the characteristics of the target categories,
	and the blue and orange borders indicate that the target category is sampled from unseen classes and seen classes,
	respectively.
	The remaining images are the translation results of our proposed method.
	}
	\label{fig:examples_cub}
\end{figure}

As shown in Fig. \ref{fig:examples_cub},
when translating an input image to multiple target categories,
our results contain the required visual characteristics while remaining other class-independent information (e.g., pose, background).
This indicates that our method achieves feature disentanglement.
By encouraging the generator to exploit the semantic attribute vectors,
the class-dependent information of the output comes from the semantic attribute vector
and the class-independent information is extracted from the input image.

\clearpage

\clearpage

\begin{table}[t]
\centering
\caption{Quantitative comparison on CUB and FLO.}
\scalebox{0.55}[0.55]{
\begin{tabular}{l|rrr|rrr|rrr|rrr}
\hline
\textbf{} & \multicolumn{6}{c|}{\textbf{CUB}}                                                                                                                                                                                                                                           & \multicolumn{6}{c}{\textbf{FLO}}                                                                                                                                                                                                                                               \\ \cline{2-13} 
          & \multicolumn{3}{c|}{\textbf{Unseen}}                                                                                                          & \multicolumn{3}{c|}{\textbf{Seen}}                                                                                                            & \multicolumn{3}{c|}{\textbf{Unseen}}                                                                                                     & \multicolumn{3}{c}{\textbf{Seen}}                                                                                                             \\ \cline{2-13} 
          & \textbf{Top-1} $\uparrow$ & \textbf{Top-5} $\uparrow$ & \textbf{FID} $\downarrow$ & \textbf{Top-1} $\uparrow$ & \textbf{Top-5} $\uparrow$ & \textbf{FID} $\downarrow$ & \textbf{Top-1} $\uparrow$ & \textbf{Top-5} $\uparrow$ & \multicolumn{1}{c|}{\textbf{FID} $\downarrow$} & \textbf{Top-1} $\uparrow$ & \textbf{Top-5} $\uparrow$ & \textbf{FID} $\downarrow$ \\ \hline
FUNIT-1   & 4.68                                       & 22.80                                      & 41.48                                      & 29.53                                      & 55.73                                      & \underline{32.33}                                      & 6.90                                       & 32.00                                      & 77.77                                 & \underline{49.17}                                      & 71.12                                      & 37.56                                      \\
FUNIT-5   & \underline{7.48}                                       & \underline{31.96}                                      & \textbf{39.09}                                      & \underline{41.59}                                      & \underline{68.92}                                      & \textbf{29.34}                                      & \underline{9.00}                                       & \underline{40.70}                                      & \underline{75.87}                                 & \textbf{64.10}                                      & \textbf{83.22}                                      & \textbf{36.27}                                      \\ \hline
StarGAN   & 4.52                                       & 17.08                                      & 142.32                                     & 6.47                                       & 20.72                                      & 136.77                                     & 2.30                                       & 12.50                                      & 75.99                                 & 3.10                                       & 12.81                                      & 40.86                                      \\
Ours      & \textbf{16.88}                                      & \textbf{57.48}                                      & \underline{40.78}                                      & \textbf{59.81}                                      & \textbf{83.49}                                      & 34.89                                      & \textbf{9.30}                                       & \textbf{40.90}                                      & \textbf{73.17}                                 & 49.07                                      & \underline{73.07}                                      & \underline{37.44}                                      \\ \hline
GT        & 82.84                                      & 95.65                                      & -                                          & 84.69                                      & 96.50                                      & -                                          & 97.59                                      & 100.00                                     & -                                     & 97.86                                      & 99.68                                      & -                                          \\ \hline
\end{tabular}
\label{table:main_results}}
\end{table}

\subsubsection{Quantitative Evaluation.}
As shown in Table \ref{table:main_results},
our proposed method achieves the best performance on all the metrics over the fair baseline and shows competitive results against FUNIT-1 and FUNIT-5 especially on unseen classes.
On the FLO dataset,
the overall performance is inferior to the performance on the CUB dataset.
This is because the text embeddings used in FLO
have less information than the semantic attributes used in CUB 
and are not sufficient to represent the visual characteristics of the category.
As the total number of real images from unseen classes in FLO is smaller than the number in CUB,
the statistics of the distribution of unseen classes are more biased,
and the FID score is higher.
As the user study shown in Table~\ref{table:user_study},
our model outperforms the zero-shot baseline and achieves significant results comparable to the few-shot methods.
It indicates that the proposed learning strategy expands the sampling space and improves the model capacity for generating unseen images.

\begin{table}[t]
\begin{minipage}[b]{.4\linewidth}
\centering
\caption{Human preference score.}
\scalebox{0.6}[0.6]{
\begin{tabular}{lrr}
\hline
\textbf{Method} & \textbf{CUB}    & \textbf{FLO}    \\ \hline
FUNIT-1         & 27.8\%          & 21.8\%          \\
FUNIT-5         & \textbf{34.2\%} & 27.8\%          \\ \hline
StarGAN         & 7.8\%           & 14.3\%          \\
Ours            & 30.2\%          & \textbf{36.1\%} \\ \hline
\end{tabular}
\label{table:user_study}}
\end{minipage}
\hfill
\begin{minipage}[b]{.6\linewidth}
\centering
\caption{Ablation study on CUB.}
\scalebox{0.6}[0.6]{
\begin{tabular}{l|ccc|ccc}
\hline
\textbf{}           & \multicolumn{3}{c|}{\textbf{Unseen}}                                                                & \multicolumn{3}{c}{\textbf{Seen}}                                                                           \\ \cline{2-7} 
                    & \textbf{Top-1} & \textbf{Top-5} & \textbf{FID}   & \textbf{Top-1} & \textbf{Top-5} & \textbf{FID}   \\ \hline
Baseline            & 7.32                            & 37.48                           & 46.35                           & 57.24                           & 81.11                           & 39.22                           \\
Baseline+Preserving & 12.36                           & 46.52                           & 42.50                           & 59.69                           & \textbf{83.92} & 38.02                           \\
Baseline+Expanding  & 15.44                           & 54.24                           & 41.81                           & 54.92                           & 79.31                           & 37.13                           \\
Ours                & \textbf{16.88} & \textbf{57.48} & \textbf{40.78} & \textbf{59.81} & 83.49                           & \textbf{34.89} \\ \hline
\end{tabular}
\label{table:ablation}}
\end{minipage}
\end{table}

\subsection{Ablation Study}

The intuitive idea of solving the zero-shot I2I is to model the mapping between the attribute space and the visual space. 
One approach is to find a shared space for visual and attribute modalities~\cite{ZstGAN},
but we find that it is difficult to learn relevant representations of spatial structures when the visual samples are missing.
In this section,
we validate the importance of the two strategies we proposed,
which are to preserve semantic relations and to expand attribute space.
We perform an ablation study with two variants of our baseline on CUB,
namely,
Baseline+Preserving and Baseline+Expanding,
respectively.

\begin{figure}[t]
	\centering
	\includegraphics[width=0.75\linewidth]{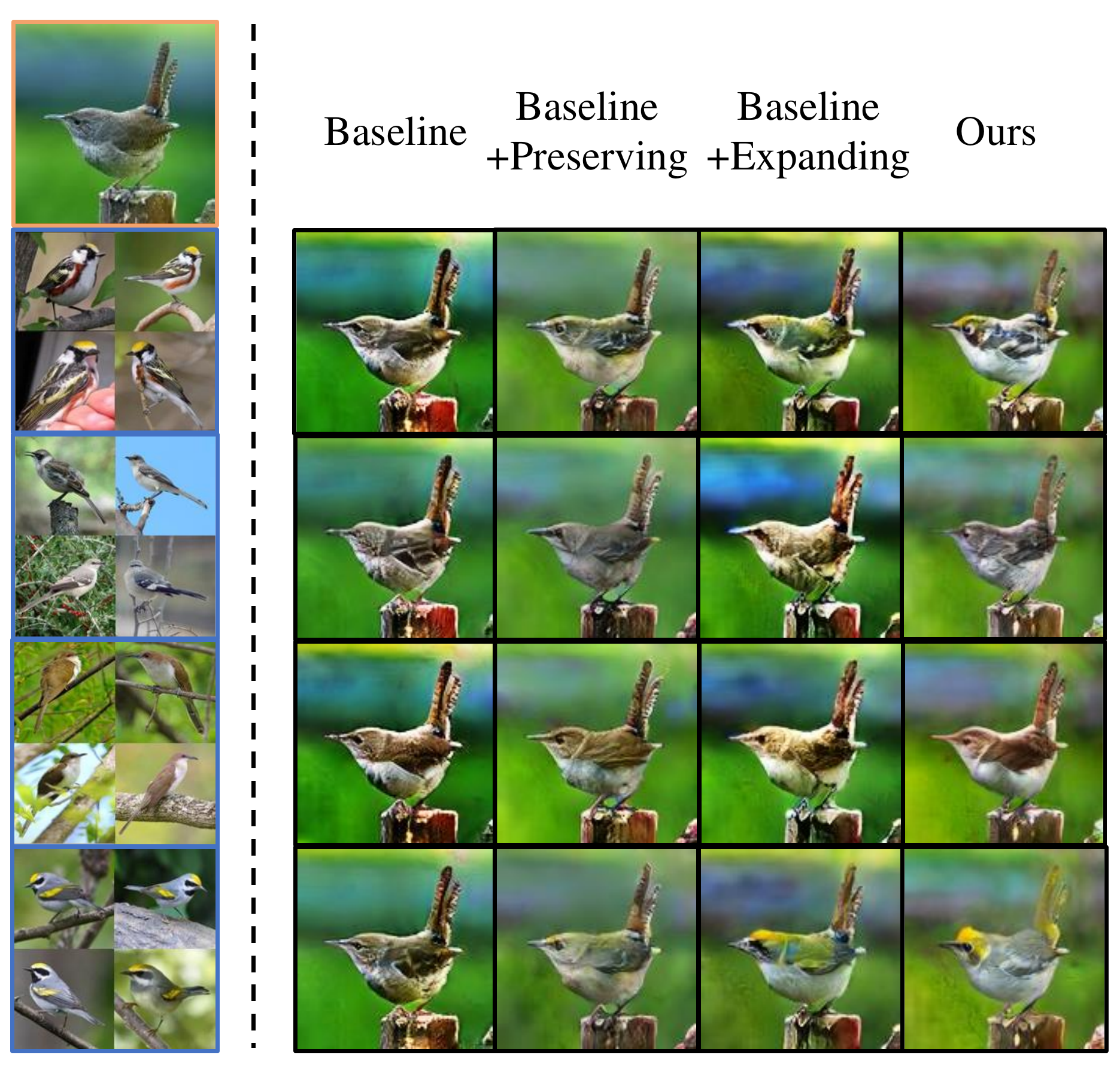}
	\caption{
	Qualitative comparison for ablation study.
	The image with an orange border is the input image. 
	and the other images in the first column represent the characteristics of the target categories which are sampled from unseen classes.
	Each of the remaining columns shows the outputs from a method.
	}
	\label{fig:ablation}
\end{figure}

The quantitative results in Table \ref{table:ablation} indicate
the effectiveness of our proposed learning strategies.
By adding either of these two strategies to the baseline model,
the performance gap between unseen classes and seen classes is narrowed.
By combining these two strategies,
our proposed method obtains the best results on almost all the metrics.
The results in Fig. \ref{fig:ablation} show the qualitative comparison for
the ablation study.
Without the proposed strategies for generalizing to unseen classes,
the baseline model suffers from the problem of mode collapse.
All of its outputs have a similar appearance 
and can not be well classified into the target categories.
In contrast,
the results of our proposed method contain the visual characteristics of the target categories.

\begin{figure}[h]
	\centering
	\includegraphics[width=0.98\linewidth]{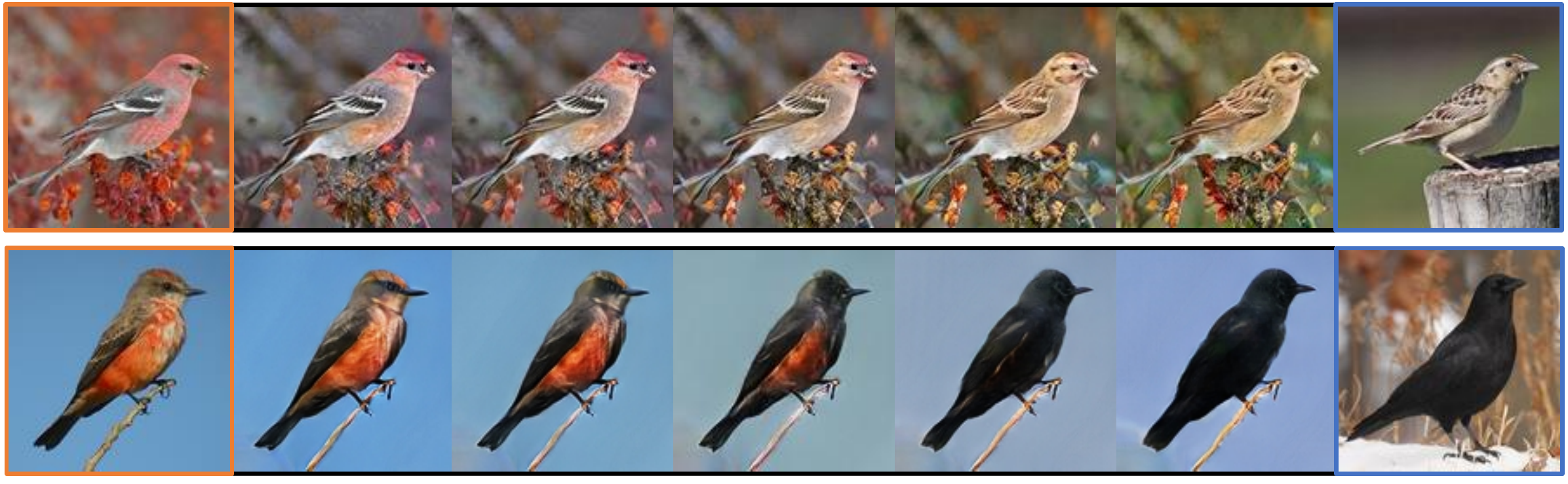}
	\caption{
	Interpolation between two attributes of seen and unseen classes.
	The images with an orange border are the input images sampled from seen classes,
	while the images with a blue border represent the characteristics of the target classes.
	}
	\label{fig:interpolation}
\end{figure}

\subsection{Attribute Interpolation}
In this section,
by linearly interpolating between two attributes of seen and unseen classes,
we visualize the intermediate results of the mapping from seen classes to unseen classes.
As shown in Fig. \ref{fig:interpolation},
the interpolation results demonstrate the continuity in the attribute space
and verify the effectiveness of our proposed two strategies for exploiting the attribute space.

\subsection{Effect of the Number of Seen Classes}

To analyze the effect of the number of seen classes,
we train our model with different dataset splits,
using the CUB dataset.
As shown in Table \ref{table:num_of_seen},
all the performance metrics are positively correlated with the number of seen classes.
The more classes available during training time,
the more visual patterns will be learned 
and these patterns will be better linked to the semantic attributes,
leading to better performance at test time.
An outlier is the FID score on seen classes when the number of seen classes is 180,
which is 64.59 and larger than others.
We conjecture that this is because when the number of seen classes becomes 180,
the number of unseen classes is reduced to 20,
thus the size of the unseen dataset is not adequate to calculate an unbiased FID score.

\subsection{Application}
In the above,
we have verified the effectiveness of our proposed strategies.
In this section,
we explore the application of our method in zero-shot classification and fashion design.

\subsubsection{Zero-Shot Classification.}
We demonstrate that the proposed zero-shot unsupervised I2I translation scheme can benefit zero-shot classification.
After the training of our proposed image translation model,
we first generate new samples for each unseen class to augment the training dataset,
thus the zero-shot classification becomes a conventional classification problem.
Then we train a classifier based on this new augmented dataset that contains samples 
for both seen and unseen classes.
Similar to the deep CNN encoder used in recent zero-shot classification methods~\cite{xian2018zero,xian2018feature},
we adopt a ResNet-101~\cite{he2016deep} as our classifier.

\begin{table}[t]
\begin{minipage}[b]{.5\linewidth}
\centering
\caption{Effect of the number of seen classes on CUB.}
\scalebox{0.6}[0.6]{
\begin{tabular}{c|ccc|ccc}
\hline
\textbf{} & \multicolumn{3}{c|}{\textbf{Unseen}}                                                                & \multicolumn{3}{c}{\textbf{Seen}}                                                                  \\ \cline{2-7} 
          & \textbf{Top-1} & \textbf{Top-5} & \textbf{FID}   & \textbf{Top-1} & \textbf{Top-5} & \textbf{FID}   \\ \hline
90        & 10.53                           & 43.09                           & 42.96                           & 56.22                           & 79.73                           & 41.84                           \\
120       & 15.85                           & 52.38                           & 42.10                           & 56.40                           & 81.22                           & 39.06                           \\
150       & 16.88                           & 57.48                           & \textbf{40.78} & 59.81                           & 83.49                           & 34.89                           \\
180       & \textbf{22.20} & \textbf{66.20} & 64.59                           & \textbf{61.26} & \textbf{85.24} & \textbf{34.53} \\ \hline
\end{tabular}
\label{table:num_of_seen}}
\end{minipage}
\hfill
\begin{minipage}[b]{.45\linewidth}
\centering
\caption{Results of zero-shot classification on CUB.}
\scalebox{0.6}[0.6]{
\begin{tabular}{l|lcc}
\hline
\textbf{} & \textbf{U} & \textbf{S}               & H                        \\ \hline
f-CLSWGAN~\cite{xian2018feature} & 43.7       & 57.7                     & 49.7                     \\
LisGAN~\cite{li2019leveraging}  & 46.5       & 57.9                     & 51.6                     \\
CADA-VAE~\cite{schonfeld2019generalized}  & \underline{51.6}       & 53.5                     & 52.4                     \\
ZSL-ABP~\cite{zhu2019learning}   & 47.0       & 54.8                     & 50.6                     \\
SABR-I~\cite{paul2019semantically}    & \textbf{55.0}       & \multicolumn{1}{l}{58.7} & \multicolumn{1}{l}{\textbf{56.8}} \\
AREN~\cite{xie2019attentive}      & 38.9       & \multicolumn{1}{l}{\textbf{78.7}} & \multicolumn{1}{l}{52.1} \\ \hline
ours      & 49.7       & \underline{65.9}                     & \underline{56.7}                     \\ \hline
\end{tabular}
\label{table:classification}}
\end{minipage}
\end{table}

We validate the classification performance under a generalized zero-shot learning(GZSL) setting.
During the test phase,
we use images from both seen and unseen classes,
and the label space is the combination of seen and unseen classes
$\mathcal{Y}^s \cup \mathcal{Y}^u$.
The task of the classifier is to learn a mapping
$ f_{GZSL} : \mathcal{X}^s \cup \mathcal{X}^u  \rightarrow \mathcal{Y}^s \cup \mathcal{Y}^u $.
The performance of the classifier is evaluated based on the harmonic mean(H) of the top-1 classification accuracy on seen and unseen classes,
i.e.,
S and U.

We compare our model with recent state-of-the-art methods,
and the results in Table \ref{table:classification} show that our proposed method is comparable to these methods on CUB.
Our model achieves the competitive result of harmonic mean accuracy,
which indicates that it obtains a balance between seen and unseen classes.
This experiment validates the effectiveness of our proposed strategies,
and exploring the use of the proposed strategies for feature level
synthesis is interesting for future work.

\subsubsection{Fashion Design.}
\begin{figure}[t]
	\centering
	\includegraphics[width=0.98\linewidth]{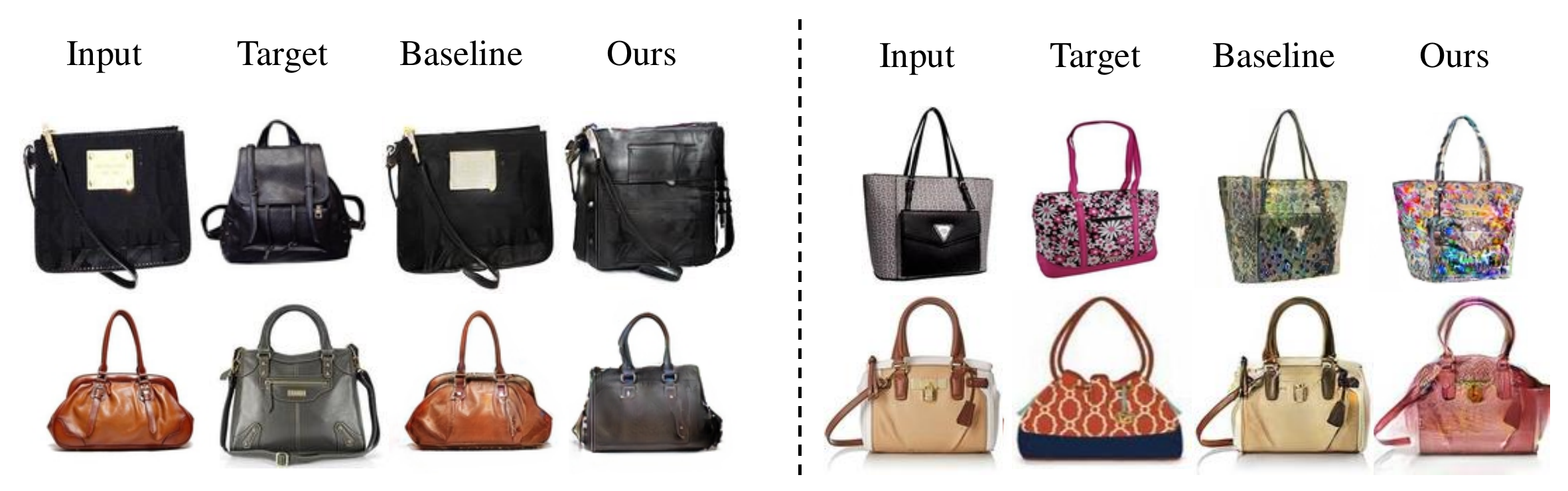}
	\caption{
	Example results of handbag translation. The input images are sampled from seen classes, while the target images are sampled from unseen classes.
	}
	\label{fig:bags}
\end{figure}

Automatic fashion design has recently attracted great attention due to its lucrative applications~\cite{liu2019toward}. 
Given the semantic description of a novel handbag,
our task is to synthesize images with the required attributes and input images.
The model is trained on the data from \cite{zhu2016generative}.
The example results shown in Fig.\ref{fig:bags} demonstrate the effectiveness of our method.
Compared with our baseline,
when translating source image to target novel classes,
the outputs produced by our method captures realistic and relevant characteristics.
That is because the baseline model lacks enough incentives to utilize the attribute vector,
while with the proposed strategies,
the model is able to bridge the semantic attributes and the visual characteristics effectively.
More details about the experiment can be found in our supplementary materials. 

\section{Conclusion} \label{sec:conclusion}

In this paper, we explore the phenomenon that the multi-domain translation model suffers from mode collapse when synthesizing images of unseen classes. To address this issue, we extend the multi-domain baseline to a zero-shot paradigm by preserving semantic relations and expanding attribute space. Experiments demonstrate that the proposed method can generate realistic images while relevant to their target categories for both unseen and seen classes. Besides, we show that our model is beneficial for many applications such as zero-shot classification and fashion design.









\section*{Acknowledgement}

This work was supported in part by Key-Area Research and Development Program of Guangdong Province (No. 2019B121204008),
Shenzhen Municipal Science and Technology Program (No. JCYJ20170818141146428),
National Natural Science Foundation of China and Guangdong Province Scientific Research on Big Data (No. U1611461).

\bibliographystyle{elsarticle-num}
\bibliography{bib}

\appendix

\section{Network Architecture}

The architecture details are shown in Table \ref{table:supp_dis},
\ref{table:supp_gen}
and \ref{table:supp_mlp}.
In addition to the main architecture,
the generator $G$ contains a 3-layer MLP that learns a set of affine transformation parameters
using the attribute $a$.
There are some notations:
$h$ and $w$: height and width of the input image,
$n_a$: the dimension of attribute $a$,
$n_c$: the number of seen classes,
N: the number of output channels,
K: kernel size,
S: stride size,
P: padding size,
FC: fully connected layer,
IN: instance normalization,
AdaIN: adaptive instance normalization,
LReLU: Leaky ReLU with a negative slope of 0.2.

\begin{table}[h]
\centering
\scalebox{0.75}[0.75]{
\begin{tabular}{@{}ccc@{}}
\toprule
\textbf{Part}   & \textbf{Input $\rightarrow$ Output Shape}       & \textbf{Layer Information}                  \\ \midrule
Input Layer     & (h,w,3) $\rightarrow$ (h,w,64)                  & CONV-(N64, K7$\times$7, S1, P3)                    \\ \midrule
                & (h,w,64) $\rightarrow$ (h,w,64)                 & ResBlock: LReLU, CONV-(N64, K3$\times$3, S1, P1)   \\ \cmidrule(l){2-3} 
                & (h,w,64) $\rightarrow$ (h,w,128)                & ResBlock: LReLU, CONV-(N128, K3$\times$3, S1, P1)  \\ \cmidrule(l){2-3} 
                & (h,w,128) $\rightarrow$ ($\frac{h}{2}$,$\frac{w}{2}$,128)           & AvgPool-(K3$\times$3, S2, P1)                      \\ \cmidrule(l){2-3} 
                & ($\frac{h}{2}$,$\frac{w}{2}$,128) $\rightarrow$ ($\frac{h}{2}$,$\frac{w}{2}$,128)       & ResBlock: LReLU, CONV-(N128, K3$\times$3, S1, P1)  \\ \cmidrule(l){2-3} 
                & ($\frac{h}{2}$,$\frac{w}{2}$,128) $\rightarrow$ ($\frac{h}{2}$,$\frac{w}{2}$,256)       & ResBlock: LReLU, CONV-(N258, K3$\times$3, S1, P1)  \\ \cmidrule(l){2-3} 
                & ($\frac{h}{2}$,$\frac{w}{2}$,256) $\rightarrow$ ($\frac{h}{4}$,$\frac{w}{4}$,256)       & AvgPool-(K3$\times$3, S2, P1)                      \\ \cmidrule(l){2-3} 
Hidden Layers    & ($\frac{h}{4}$,$\frac{w}{4}$,256) $\rightarrow$ ($\frac{h}{4}$,$\frac{w}{4}$,256)       & ResBlock: LReLU, CONV-(N256, K3$\times$3, S1, P1)  \\ \cmidrule(l){2-3} 
                & ($\frac{h}{4}$,$\frac{w}{4}$,256) $\rightarrow$ ($\frac{h}{4}$,$\frac{w}{4}$,512)       & ResBlock: LReLU, CONV-(N512, K3$\times$3, S1, P1)  \\ \cmidrule(l){2-3} 
                & ($\frac{h}{4}$,$\frac{w}{4}$,512) $\rightarrow$ ($\frac{h}{8}$,$\frac{w}{8}$,512)       & AvgPool-(K3$\times$3, S2, P1)                      \\ \cmidrule(l){2-3} 
                & ($\frac{h}{8}$,$\frac{w}{8}$,512) $\rightarrow$ ($\frac{h}{8}$,$\frac{w}{8}$,512)       & ResBlock: LReLU, CONV-(N512, K3$\times$3, S1, P1)  \\ \cmidrule(l){2-3} 
                & ($\frac{h}{8}$,$\frac{w}{8}$,512) $\rightarrow$ ($\frac{h}{8}$,$\frac{w}{8}$,1024)      & ResBlock: LReLU, CONV-(N1024, K3$\times$3, S1, P1) \\ \cmidrule(l){2-3} 
                & ($\frac{h}{8}$,$\frac{w}{8}$,1024) $\rightarrow$ ($\frac{h}{16}$,$\frac{w}{16}$,1024)   & AvgPool-(K3$\times$3, S2, P1)                      \\ \cmidrule(l){2-3} 
                & ($\frac{h}{16}$,$\frac{w}{16}$,1024) $\rightarrow$ ($\frac{h}{16}$,$\frac{w}{16}$,1024) & ResBlock: LReLU, CONV-(N1024, K3$\times$3, S1, P1) \\ \cmidrule(l){2-3} 
                & ($\frac{h}{16}$,$\frac{w}{16}$,1024) $\rightarrow$ ($\frac{h}{16}$,$\frac{w}{16}$,1024) & ResBlock: LReLU, CONV-(N1024, K3$\times$3, S1, P1) \\ \midrule
Output Layer(D) & ($\frac{h}{16}$,$\frac{w}{16}$,1024) $\rightarrow$ ($\frac{h}{16}$,$\frac{w}{16}$,$n_c$)   & LReLU, CONV-(N($n_c$), K1$\times$1, S1, P0)           \\ \midrule
                & ($\frac{h}{16}$,$\frac{w}{16}$,1024) $\rightarrow$ ($\frac{h}{16}$,$\frac{w}{16}$,$n_a$)   & LReLU, CONV-(N($n_a$), K1$\times$1, S1, P0)           \\ \cmidrule(l){2-3} 
Output Layer(R) & ($\frac{h}{16}$,$\frac{w}{16}$,$n_a$) $\rightarrow$ ($n_a$)               & LReLU, GlobalAvgPool                        \\ \cmidrule(l){2-3} 
                & ($n_a$) $\rightarrow$ ($n_a$)                         & FC-($n_a$, $n_a$), Sigmoid                        \\ \bottomrule
\end{tabular}}
\label{table:supp_dis}
\caption{Architecture of the discriminator.}
\end{table}

\clearpage

\begin{table}[h]
\centering
\scalebox{0.7}[0.7]{
\begin{tabular}{@{}ccc@{}}
\toprule
\textbf{Part}        & \textbf{Input $\rightarrow$ Output Shape}          & \textbf{Layer Information}                       \\ \midrule
Input Layer          & (h,w,3) $\rightarrow$ (h,w,64)                     & CONV-(N64, K7$\times$7, S1, P3), IN, ReLU               \\ \midrule
                     & (h,w,64) $\rightarrow$ ($\frac{h}{2}$,$\frac{w}{2}$,128)               & CONV-(N128, K4$\times$4, S2, P1), IN, ReLU              \\ \cmidrule(l){2-3} 
Down-sampling        & ($\frac{h}{2}$,$\frac{w}{2}$,128) $\rightarrow$ ($\frac{h}{4}$,$\frac{w}{4}$,256)          & CONV-(N256, K4$\times$4, S2, P1), IN, ReLU              \\ \cmidrule(l){2-3} 
                     & ($\frac{h}{4}$,$\frac{w}{4}$,256) $\rightarrow$ ($\frac{h}{8}$,$\frac{w}{8}$,512)          & CONV-(N512, K4$\times$4, S2, P1), IN, ReLU              \\ \midrule
                     & ($\frac{h}{8}$,$\frac{w}{8}$,512) $\rightarrow$ ($\frac{h}{8}$,$\frac{w}{8}$,512)          & ResBlock: CONV-(N128, K3$\times$3, S1, P1), IN, ReLU    \\ \cmidrule(l){2-3} 
Bottleneck           & ($\frac{h}{8}$,$\frac{w}{8}$,512) $\rightarrow$ ($\frac{h}{8}$,$\frac{w}{8}$,512)          & ResBlock: CONV-(N128, K3$\times$3, S1, P1), IN, ReLU    \\ \cmidrule(l){2-3} 
\multicolumn{1}{l}{} & ($\frac{h}{8}$,$\frac{w}{8}$,512) + (2048) $\rightarrow$ ($\frac{h}{8}$,$\frac{w}{8}$,512) & ResBlock: CONV-(N128, K3$\times$3, S1, P1), AdaIN, ReLU \\ \cmidrule(l){2-3} 
                     & ($\frac{h}{8}$,$\frac{w}{8}$,512) + (2048) $\rightarrow$ ($\frac{h}{8}$,$\frac{w}{8}$,512) & ResBlock: CONV-(N128, K3$\times$3, S1, P1), AdaIN, ReLU \\ \midrule
                     & ($\frac{h}{8}$,$\frac{w}{8}$,512) $\rightarrow$ ($\frac{h}{4}$,$\frac{w}{4}$,256)          & Upsample(2), CONV-(N256, K5$\times$5, S1, P2), IN, ReLU \\ \cmidrule(l){2-3} 
Up-sampling          & ($\frac{h}{4}$,$\frac{w}{4}$,256) $\rightarrow$ ($\frac{h}{2}$,$\frac{w}{2}$,128)          & Upsample(2), CONV-(N128, K5$\times$5, S1, P2), IN, ReLU \\ \cmidrule(l){2-3} 
                     & ($\frac{h}{2}$,$\frac{w}{2}$,128) $\rightarrow$ (h,w,64)               & Upsample(2), CONV-(N64, K5$\times$5, S1, P2), IN, ReLU  \\ \midrule
Output Layer         & (h,w,64) $\rightarrow$ (h,w,3)                     & CONV-(N3, K7$\times$7, S1, P3), Tanh                    \\ \bottomrule
\end{tabular}}
\caption{Main architecture of the generator.}
\label{table:supp_gen}
\end{table}

\begin{table}[hb]
\centering
\scalebox{0.7}[0.7]{
\begin{tabular}{@{}ccc@{}}
\toprule
\textbf{Part} & \textbf{Input $\rightarrow$ Output Shape} & \textbf{Layer Information} \\ \midrule
Input Layer   & ($n_a$) $\rightarrow$ (256)                  & FC-($n_a$, 256), ReLU         \\ \midrule
Hidden Layer  & (256) $\rightarrow$ (256)                 & FC-(256, 256), ReLU        \\ \midrule
Output Layer  & (256) $\rightarrow$ (4096)                & FC-(256, 4096)             \\ \bottomrule
\end{tabular}}
\caption{Architecture of the MLP in the generator.}
\label{table:supp_mlp}
\end{table}

\section{Experiment of Fashion Design}
The dataset is from \cite{zhu2016generative}.
We randomly select a subset of it,
which contains 10,000 images of various handbags.
We adopt an Inception-V3~\cite{szegedy2016rethinking} network pre-trained on the ImageNet~\cite{russakovsky2015imagenet} dataset to extract embeddings of these images.
The 2048-dim embeddings serve as semantic descriptions of images.
We then run K-means clustering on the embeddings of all images and get 50 classes.
The dataset is randomly split into 40 training classes and 10 test classes.

\begin{figure}[h]
	\centering
	\includegraphics[width=0.98\linewidth]{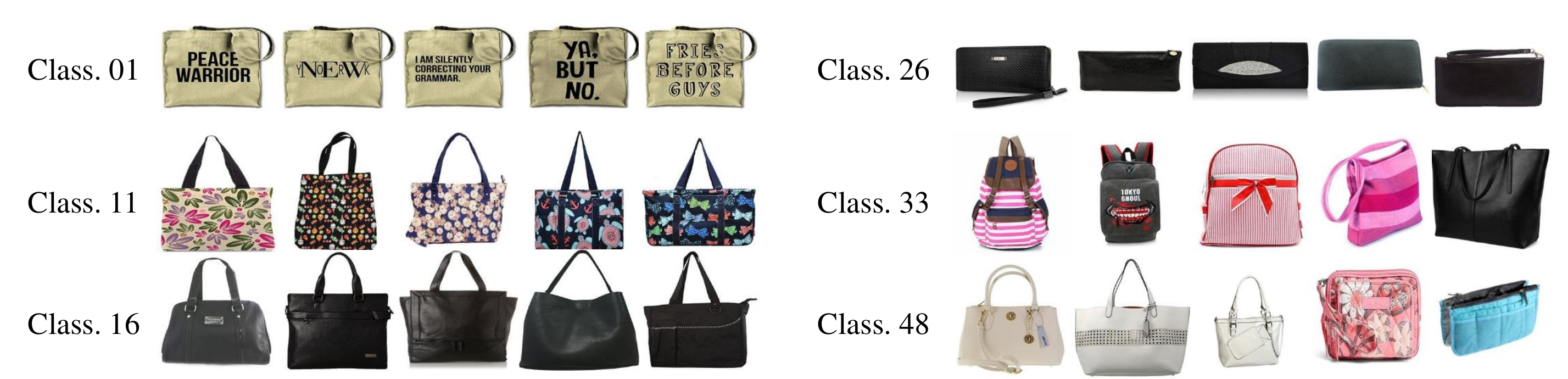}
	\caption{Examples of the clustering results.
	}
	\label{fig:supp_bags}
\end{figure}

Fig.~\ref{fig:supp_bags} shows some examples of the clustering results.
Most images have consistent characteristics within classes.
However,
a few classes, such as Class.~33 and Class.~48, are not clustered well and have various patterns,
which degrades the performance of the zero-shot image-to-image translation.
We conjecture that this is because our embeddings are not sufficient to represent the visual characteristics of the handbag images.
Fine-tuning the Inception-V3 on the handbag dataset or finding more suitable semantic descriptions for this application may help address this issue and boost the performance.

\section{Additional Qualitative Results}

\begin{figure}[h]
	\centering
	\includegraphics[width=0.7\linewidth]{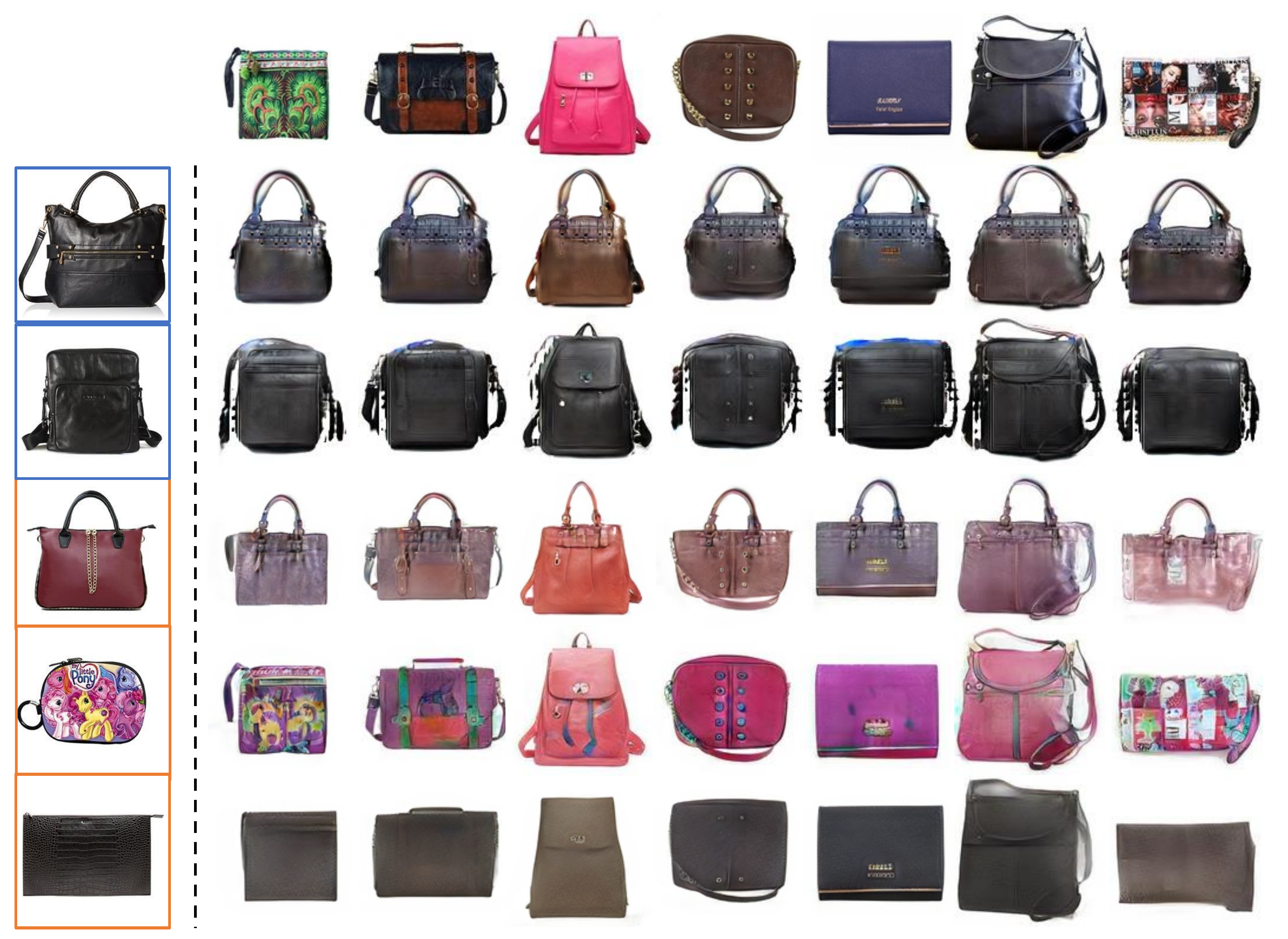}
	\caption{
	\textbf{Qualitative results of fashion design.} 
	The first row shows the input images sampled from seen classes.
	The first column represents the characteristics of the target categories,
	and the blue and orange borders indicate that the target category is sampled from unseen classes and seen classes,
	respectively.
	The remaining images are the translation results of our proposed method.
	}
	\label{fig:supp_bags2}
\end{figure}

\begin{figure}[t]
	\centering
	\includegraphics[width=0.95\linewidth]{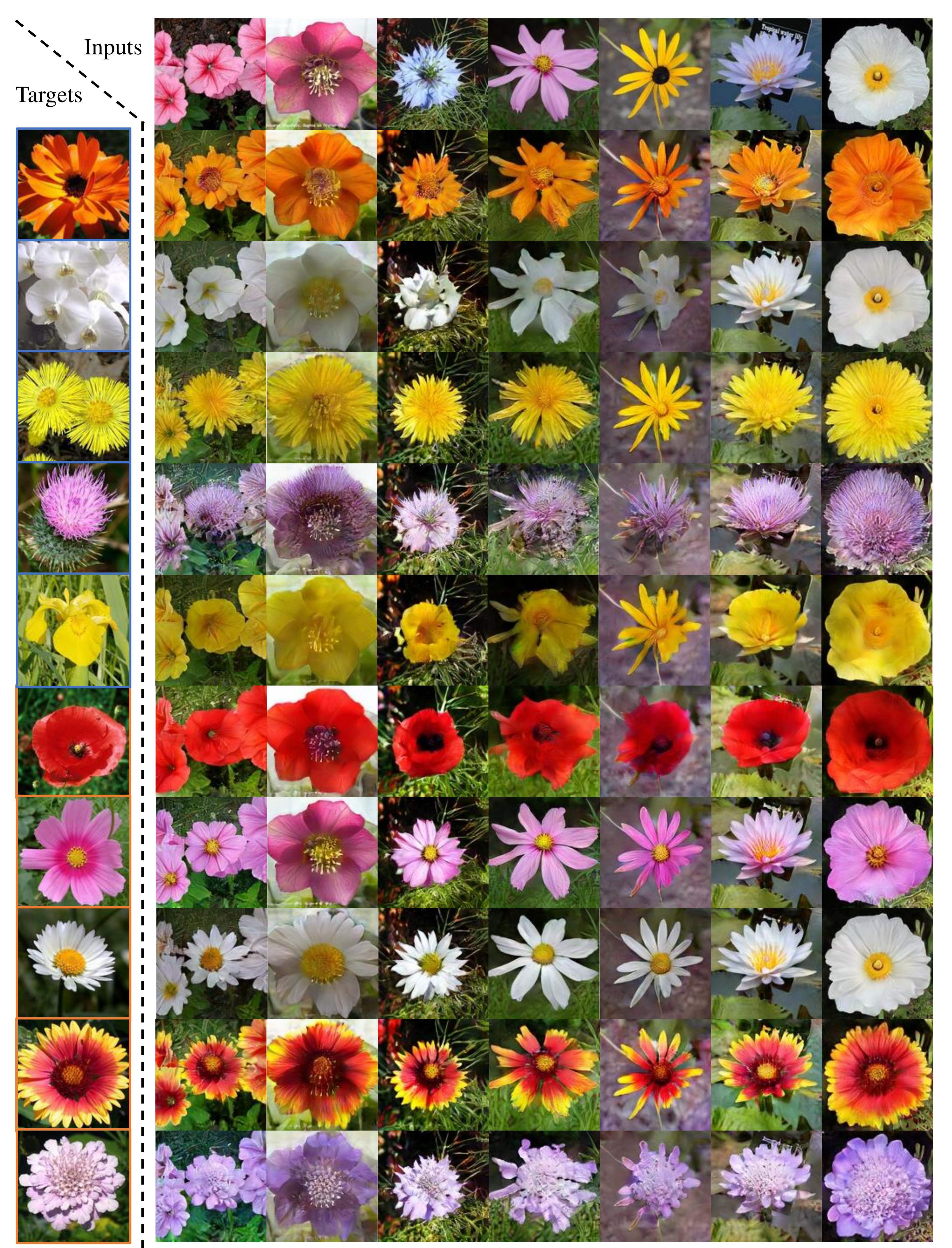}
	\caption{
	\textbf{Qualitative results on FLO.} 
	The first row shows the input images sampled from seen classes.
	The first column represents the characteristics of the target categories,
	and the blue and orange borders indicate that the target category is sampled from unseen classes and seen classes,
	respectively.
	The remaining images are the translation results of our proposed method.
	}
	\label{fig:supp_flo}
\end{figure}

\clearpage

\end{document}